\documentclass{article}

\usepackage[preprint]{neurips_2025}

\usepackage[utf8]{inputenc} %
\usepackage[T1]{fontenc}    %
\usepackage{hyperref}       %
\usepackage{url}            %
\usepackage{booktabs}       %
\usepackage{amsfonts}       %
\usepackage{nicefrac}       %
\usepackage{microtype}      %
\usepackage{xcolor}         %
\usepackage[pdftex]{graphicx}
\usepackage{amsmath}
\usepackage{subcaption}
\usepackage{wrapfig}

\newcommand\ourmethod{3D24D}

\title{Bringing Objects to Life: training-free 4D generation from 3D objects through view consistent noise}

\author{Ohad Rahamim$^{1}$ \hspace{0.05\linewidth} \and
Ori Malca$^{1}$ \hspace{0.05\linewidth} \and
Dvir Samuel$^{1}$ \hspace{0.05\linewidth} \and 
Gal Chechik$^{1,2}$ \and
\hspace{0.06\linewidth}  \and \hspace{0.9\linewidth}
\and \hspace{0.1\linewidth}$^{1}$Bar-Ilan University \and $^{2}$NVIDIA \and \hspace{0.05\linewidth} 
}

\begin{document}

\maketitle

\begin{figure*}[h!]
    \centering
    \includegraphics[width=\textwidth]{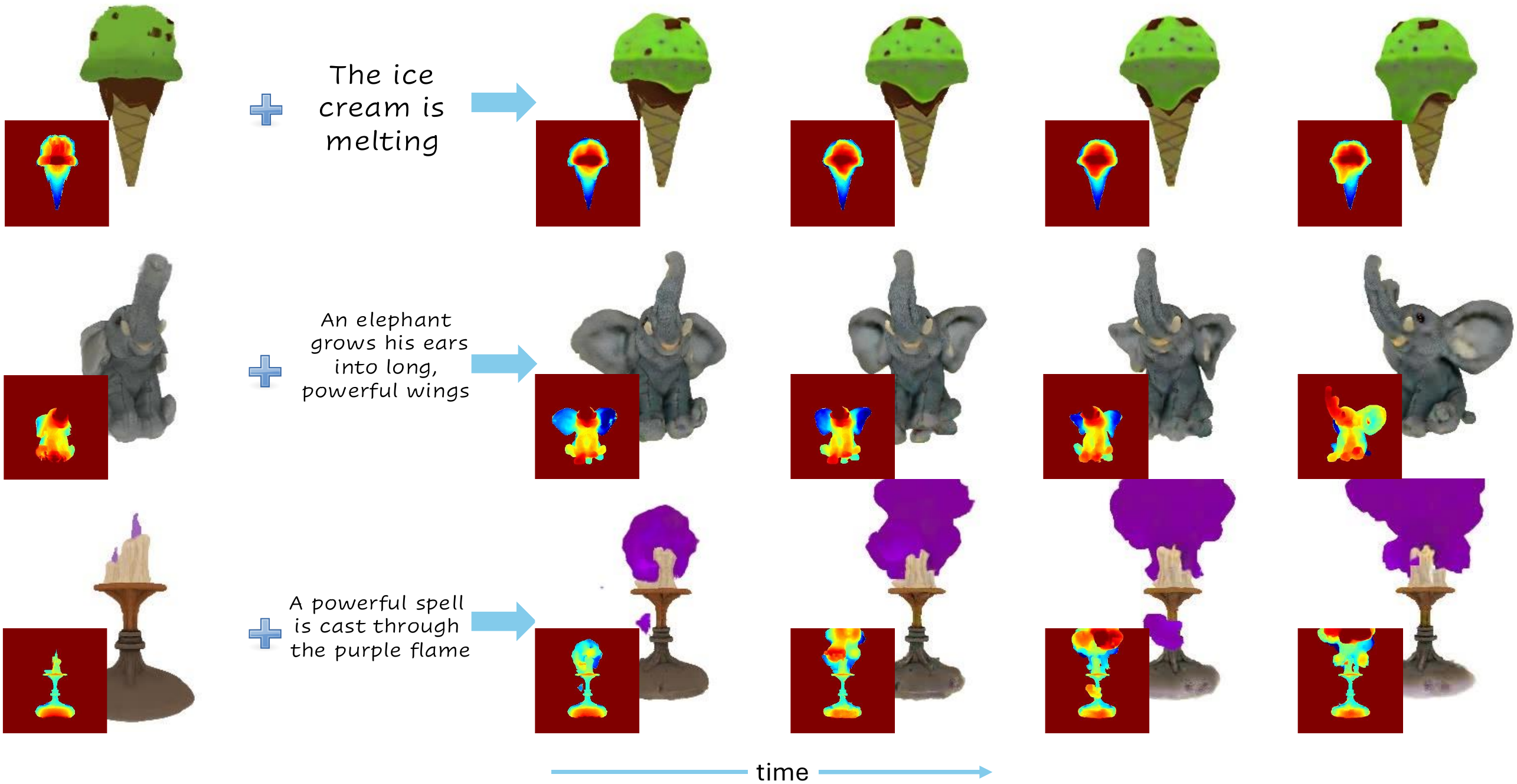}
    \caption{Our method, \ourmethod{}, takes a static 3D object and a textual prompt describing a desired action. It then adds dynamics to the object based on the prompt to create a 4D animation, essentially a video viewable from any perspective. On the right, we display four 3D frames from the generated 4D animation. Each 3D frame contains an RGB image and a corresponding depth map on its bottom left. } 
    \label{fig:teaser}
\end{figure*}

\begin{abstract}
Recent advancements in generative models have enabled the creation of dynamic 4D content — 3D objects in motion — based on text prompts, which holds potential for applications in virtual worlds, media, and gaming. Existing methods provide control over the appearance of generated content, including the ability to animate 3D objects.
However, their ability to generate dynamics is limited to the mesh datasets they were trained on, lacking any growth or structural development capability.
In this work, we introduce a training-free method for animating 3D objects by conditioning on textual prompts to guide 4D generation, enabling custom general scenes while maintaining the original object's identity. We first convert a 3D mesh into a static 4D Neural Radiance Field (NeRF) that preserves the object’s visual attributes. Then, we animate the object using an Image-to-Video diffusion model driven by text. To improve motion realism, we introduce a view-consistent noising protocol that aligns object perspectives with the noising process to promote lifelike movement, and a masked Score Distillation Sampling (SDS) loss that leverages attention maps to focus optimization on relevant regions, better preserving the original object. We evaluate our model on two different 3D object datasets for temporal coherence, prompt adherence, and visual fidelity, and find that our method outperforms the baseline based on multiview training, achieving better consistency with the textual prompt in hard scenarios. \href{https://three24d.github.io/three24d/}{Project page}
\end{abstract}

\section{Introduction}
\label{sec:intro}
Generative models are progressing rapidly, making it possible to generate images, videos, 3D objects, and scenes from text instructions only. It is now becoming possible to generate 4D content: dynamic 3D content conditioned on text prompts using text-to-4D methods~\cite{mav3d, 4dfy, aligngaussians, pla4d, 4dynamic, comp4d, tc4d, stp4d, trans4d},  
4D generation has the potential to change content creation, from movies and games to simulating virtual worlds. 

Despite this promise, text-to-4D methods provide very limited control over the appearance of generated 4D content. Instead of generating a 4D dynamic object using text control only,
latest work on 4D generation established better conditioning like image-to-4d \cite{animate124, dreamgaussian4d, gaussianflow, 4dgen} and video-to-4D \cite{sc4d, 4diffusion, sv4d, l4gm, zero4d, yang2025not}, where videos provide more information relevant to the dynamics.
The latest advancement in conditioning is 3D-to-4D generation, specifically \textit{Animate3D}~\cite{animate3d} and \textit{Diffusion4D}~\cite{diffusion4d}, which train multi-view image-to-video diffusion models. These works capture a 3D object from multiple viewpoints and generate temporally consistent videos for each, ensuring coherence across different perspectives.
To achieve this consistency, they rely on large-scale datasets of multi-view videos derived from existing 4D objects. However, these 4D objects are represented as meshes, which are inherently constrained by their fixed number of vertices and faces. As a result, approaches trained on this dataset tend to be more limited in handling evolution, volume change or growth deformation.
Moreover, training-based approaches need to be retrained for new models, which may reduce usability.

In this paper, we introduce a novel training-free method for generating 4D scenes from user-provided 3D representations, taking a simple approach that incorporates textual descriptions to govern the animation of the 3D objects.
First, we train a ``static`` 4D Neural Radiance Field (NeRF) based on the 3D mesh input, effectively capturing the object structure and appearance from multiple views, replicated across time. Then, our method modifies the 4D object using an image-to-video diffusion model \citep{dynamicrafter, I2VGen, imagenvideo, ltxv}, conditioning the first frame on renderings of the input object. This maintains the identity of the original object and adds motion based on a provided text prompt. 

Unfortunately, we find that applying this approach naively is insufficient because it dramatically reduces the level of dynamic motion. We propose two key improvements that both enhance the generation of dynamic movements and ensure better preservation of the input object.
First, we design a new view-consistent noising strategy for 4D generation, which constructs a noise pattern associated with the rendered viewpoint during optimization.
This association between the viewpoint and the noising approach enhances the generation process, resulting in more pronounced motion in the animated 4D output.
Second, we introduce a masked variant of the SDS loss that uses attention maps obtained from the image-to-video model. This masked SDS focuses optimization on the object across temporally relevant regions of the latent space, enhancing the fidelity of object-related elements and better preserving its identity.
We name our approach simply \ourmethod{}.

We evaluate \ourmethod{} on two different datasets, with a comprehensive set of metrics designed to assess various aspects of the generated 4D scenes across multiple viewpoints. We focus on four main criteria: temporal coherence of the generated video, adherence to the prompt description, and visual consistency with the initial 3D object. Given that only one 3D-to-4D generation method is publicly available \cite{animate3d}, our comparison demonstrates that our approach achieves a better alignment with the text prompt while exhibiting more relevant dynamics, for prompts that elicit significant non-rigid deformations. We also find that our proposed view-consistency noising protocol and the attention-masked SDS enhance the dynamic content of the generated videos while still maintaining a high degree of consistency with the original object's appearance. These improvements demonstrate that our method generates more realistic 4D scenes and also effectively balances visual quality and dynamic richness.
\newpage
\noindent This paper makes the following contributions:
\begin{enumerate}
    \item A novel training-free workflow for generating 4D scenes conditioned on a given 3D object model and on a text prompt.
    \item We introduce two enhancements to improve motion generation and optimization: 
        (a) A viewpoint-consistent noising strategy that aligns the noise injection process with the rendered viewpoint, creating more dynamic and coherent movement in the 4D scene.
        (b) A masked-SDS loss that uses the cross-attention mechanism of the diffusion model to enhance the optimization of 4D content.
    \item Our method reaches improved 4D quality over current baselines on an extensive set of
    metrics, for objects that undergo significant non-rigid deformations.
\end{enumerate}

\begin{figure}[t]
    \centering
    \includegraphics[width=\textwidth]{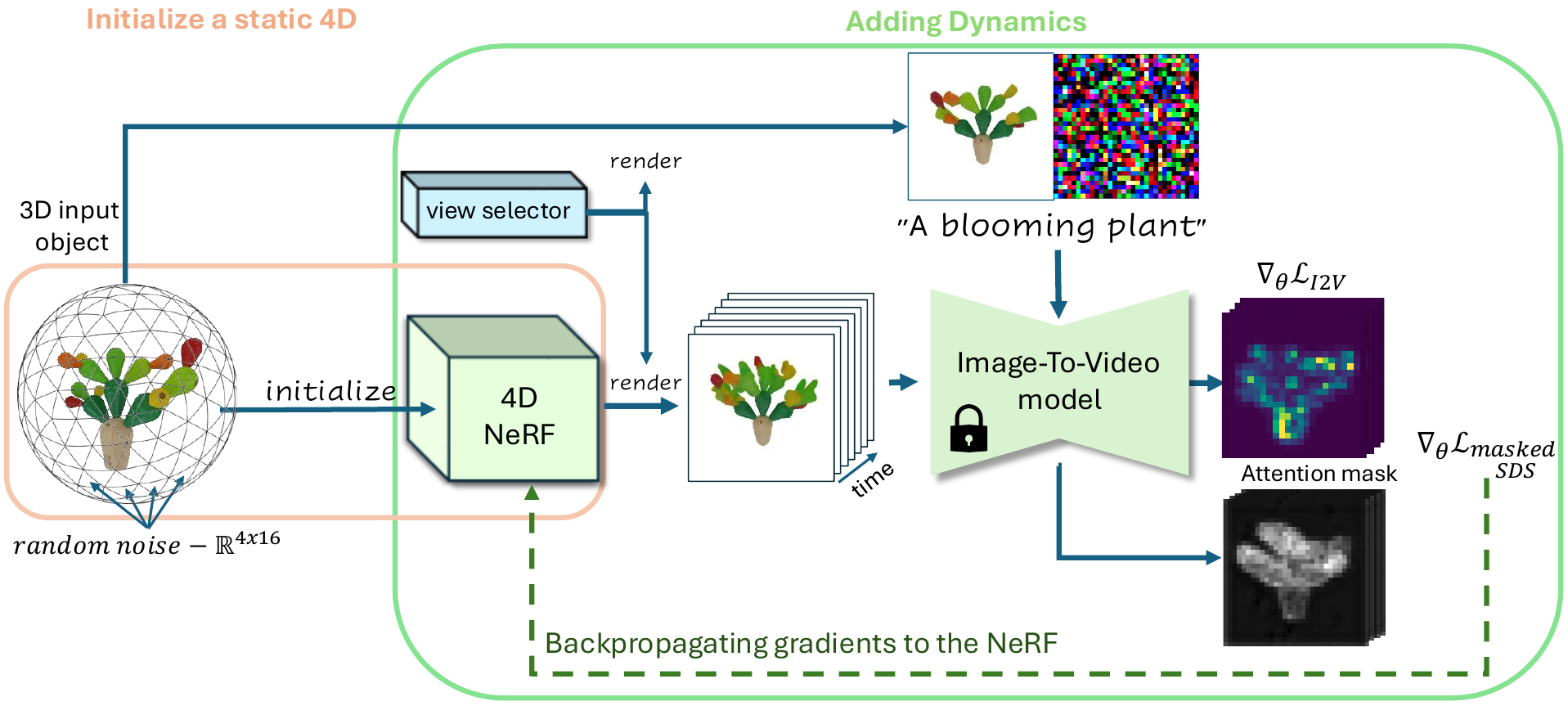}
    \caption{
    Workflow of our \ourmethod{} approach, designed to optimize a 4D radiance field using a neural representation that captures both static and dynamic elements. First, a 4D NeRF is trained to represent the static object (plant, left), having the same 3D structure at each time step. Then, we introduce dynamics to the 4D NeRF by distilling the prior from a pre-trained image-to-video model. At each SDS step, we select a viewpoint and render both the input object, the noise sphere, and the 4D NeRF from the same selected viewpoint. These renders, along with the textual prompts, are then fed into the image-to-video model, and the SDS loss is calculated to guide the generation of motion while preserving the object's identity.
    The noise is rendered from the sphere using the same viewpoint as the static object, providing better consistency at each step.
    }
    \label{fig:pipeline}
\end{figure}

\section{Related Work}
\label{Related Work}

\noindent\paragraph{4D Generation.}
Recent advances in 4D generation span several domains, reflecting the diverse ways in which temporal and spatial information can be synthesized or manipulated. In text-to-4D approaches exemplified by earlier works \cite{4dfy, aligngaussians, pla4d, 4dynamic, comp4d, tc4d, 4real}, researchers focus on converting textual prompts into dynamic 3D scenes over time, leveraging techniques like diffusion-based models and Gaussian priors to ensure coherent spatiotemporal structure. Image-to-4D methods \cite{animate124, dreamgaussian4d, gaussianflow, twosquared, 4dgen, in24d, 4k4dgen, dimensionx}, typically transform single or multiple 2D images into volumetric sequences, often employing flow estimation or learned shape priors to extrapolate consistent motion in 3D. In the video-to-4D \cite{sc4d, 4diffusion, sv4d, sv4d2, l4gm, dreamscene4d, dreammesh4d} approaches expand existing 2D video into time-varying 3D representations, introducing techniques for multi-view consistency and temporal alignment. Finally, 3D-to-4D works \cite{animate3d, diffusion4d} tackle the challenge of adding a temporal dimension to static 3D models, enabling dynamic animations or evolutions of geometry through learned or procedural transformations. Collectively, these methods highlight a rapidly evolving field aiming to bridge the gap between static 3D content and rich, time-aware volumetric experiences.
Recent advancements in 3D-to-4D generation involve training multi-view video diffusion models on collected datasets of dynamic 3D assets. Diffusion4D \cite{diffusion4d} focuses on efficient and spatially-temporally consistent 4D content generation by adapting video diffusion models trained on such datasets. Animate3D \cite{animate3d} trains a multi-view video diffusion model conditioned on multi-view renderings of a static 3D object, It introduces a spatio-temporal attention module to enhance spatial and temporal consistency.

\noindent\paragraph{3D consistent noising.}
3D consistent noising in diffusion models addresses the challenge of generating coherent 3D content from 2D diffusion models by ensuring consistency across multiple views or time. A key approach involves generating noise directly in 3D space \cite{equivdm}, such as attaching Gaussian noise as textures to 3D meshes and rendering them from different viewpoints to provide consistent noise input. This method leverages the equivariance properties of diffusion models trained with temporally consistent noise to produce video frames that align with the underlying 3D geometry, enhancing consistency in applications like video generation and scene editing. Consistent Flow Distillation \cite{CFD} also proposes applying multi-view consistent Gaussian noise directly to the underlying 3D object representation for text-to-3D generation.

\noindent\paragraph{Image to video generation.}
Image-to-video models \citep{stableVideo} condition on an image alone, whereas text-to-video models condition on a textual prompt alone. The image-to-video approach allows users to create motion directly from the provided image, whereas text-to-video models generate motion from a textual prompt, limiting the user’s ability to explicitly control the motion dynamics. Notable works that incorporate both image and prompt conditioning include I2VGen \citep{I2VGen}, which can generate high-resolution videos, and DynamiCrafter \citep{dynamicrafter}, a family of models designed to handle various input image resolutions. Both models project the input image into a text-aligned representation space using a pre-trained CLIP image encoder, similar to how text prompts are encoded. Newer approaches, such as \cite{ltxv}, employ a highly compressed Video-VAE to enable real-time, high-quality image-to-video generation.

\section{Method}
\label{Method}

Our method receives an input 3D model (like a model of your favorite plant),  and a textual prompt (like ``A plant blooming"). Our goal is to animate the object, generating a 4D scene that reflects the described action in the prompt, yielding a 4D object of your favorite flower blooming. 
This approach transforms static assets into animated objects, adding life to 3D objects by introducing motion that aligns with the user’s descriptions. Our approach is illustrated in Figure \ref{fig:pipeline}.

\subsection{Initialize a static 4D from a 3D object}
\label{conversion to nerf}
We first optimize a ``static" 4D representation, where at every time $t$, the 4D NeRF captures the same static form of the input object.
More specifically, beginning with the input 3D mesh, we randomly select a camera position. A ray is cast from the camera center through both the mesh and the neural representation. Along this ray, 3D points are sampled, and three properties are computed: color (RGB), depth, and surface normals from both representations. We then optimize the neural representation to align with the properties of the input object. This process is illustrated in the loss function:
\begin{align}
\label{static loss}
    \notag
    \mathcal{L}_{static} &= \mathcal{L}_{MAE}(RGB_{mesh}, RGB_{NeRF}) \\ \notag
    &+ \mathcal{L}_{MAE}(Depth_{mesh}, Depth_{NeRF}) \\
    &+ \mathcal{L}_{MAE}(Normal_{mesh}, Normal_{NeRF}).
\end{align}
Here, $\mathcal{L}_{MAE}$ represents the mean absolute error between the properties of the mesh and those of the neural representation.
Along this process, we also randomly sample the time dimension, resulting in a static 4D scene where the object remains unchanged over time.

\subsection{Adding Dynamics}
\label{I2V}

Next, we aim to ``bring our object to life" by introducing motion to the static 3D input object. To do this, we need to condition the SDS process on the input 3D object to achieve the desired 4D output. 
Here, we propose using image-to-video diffusion models to enhance this process. By conditioning the generation on the provided object, we align the 3D model's render from the same viewpoint as the NeRF and use this render as input to the generation model, effectively anchoring the generated motion to the object's identity.
In our proposed SDS approach, renderings of the input object condition the distillation process, guiding the generation toward both the intended object appearance and desired dynamics.
By rendering the object from all viewpoints, we can maintain the input object's identity in the 3D space while introducing motion. 
To further ensure that the object remains consistent throughout the animation, we will also use multi-view loss to preserve its characteristics across different perspectives.

Optimizing the dynamic of the 4D scene using an image-to-video model can then be done using SDS~\citep{dreamfusion} loss: 
\begin{equation}
\label{SDS_loss_i2v}
    \nabla_{\theta} \mathcal{L}_{I2V} = \mathbb{E}_{t_d, \epsilon} \left[ \omega(t_d) \left( \epsilon_{\phi} \left( z_{t_d}; t_d, y, \mathbf{X^{obj}} \right) - \epsilon \right) \frac{\partial X_{\theta}}{\partial \theta} \right]
\end{equation}
Here, $\epsilon_{\phi}$ and $\epsilon$ denote the predicted and actual noise for each video frame, respectively. We denote $X_{\theta}$ as a collection of $V$ video frames, where $X_{\theta} = \left[ x_{\theta}^0, \ldots, x_{\theta}^{V-1} \right]$, which are rendered from the representation. Additionally, the rendered object is denoted as $X^{obj}$.

This SDS loss (Eq. \ref{SDS_loss_i2v}) is then added to the static loss (Eq. \ref{static loss}), which is applied to the frame. This combined approach generates dynamic motion while ensuring that the 3D object remains consistent at $t=0$.
The overall loss is: 
\begin{equation}
\label{total loss}
    \mathcal{L} = \mathcal{L}_{I2V}(x_{\theta}^{0,...,V}) + \mathcal{L}_{MV}(x_{\theta}^{i}) +\lambda\mathcal{L}_{static}(x_{\theta}^0).
\end{equation}
Here, $\lambda$ is a weighting hyperparameter used to balance the magnitude of $\mathcal{L}_{static}$ with that of $\mathcal{L}_{I2V}$, and $x_{\theta}^0$ is the render at time $t=0$.

\subsection{Viewpoint consistent Noising}
\label{sec:faces_noise}
When computing distillation scores from text-to-image or text-to-video models, it is common practice to randomly sample both a camera position and a noise pattern at each iteration. This means that noise patterns differ across camera viewpoints.

Our key observation is that this random sampling of noise patterns across views may reduce the consistency of appearances and motions guided by the cleaning process from different views. Trying to generate a 4D object consistent with several different motions may cause optimization to converge to a less-dynamic solution. 
Indeed, we observe this degradation of motion quality.  
See ablation Sec. \ref{Results}, Table \ref{tab:ablation}, and Figure \ref{fig:ablation} for more details.

We propose a \textbf{viewpoint-consistent noise strategy} that conditions the noise on both the rendered viewpoint $s=(\theta_s, \phi_s)$ and a set of sampled time steps  $T = \{ t_{i} \mid i = 0, \dots, V-1 \}$, where $t_i \in [0 ,1]$ and $t_{i+1} > t_i$. In standard video SDS, a viewpoint $s$ and time steps $T$ are randomly selected, and Gaussian noise $\epsilon \sim \mathcal{N}(0, I)$ is independently applied to any viewpoint. This approach neglects spatial and temporal structure, often resulting in incoherent motion dynamics.

To address this, we introduce a viewpoint consistent noising mechanism $\mathcal{N}(s, T) \in \mathbb{R}^{C \times V \times H \times W}$ that varies smoothly with both viewpoint and time, where $H, W \in \mathbb{Z}$ the space dimension $C \in \mathbb{Z}$ is the features dimention and $V \in \mathbb{Z}$ is the amount of frames. We construct this by associating a canonical 3D sphere mesh with gaussian noise attributes. Each face $f$ of the sphere is assigned a latent noise vector $\mathbf{n}_f \in \mathbb{R}^{V \times C}$. Therefore, for a given viewpoint $s$, we render the sphere to obtain a pixel-to-face mapping, allowing us to construct a noise field $\mathbf{S}^{(s)} \in \mathbb{R}^{H \times W \times V \times C}$ for each predefined time anchor $t^q_i = \frac{i}{V}$, where $i = 0, \dots, V-1$.

To support arbitrary smooth time sampling, we first randomly initialize a constant latent noise tensor $\hat{S} \in \mathbb{R}^{H \times W \times V \times C}$ at the start of optimization. This tensor remains fixed throughout optimization and is used to interpolate the noise fields at each sampled time step $\hat{t} \in T$ via:
\begin{equation}
    \mathcal{N}(s, \hat{t}) = \sqrt{1 - \tau_i} \cdot \mathbf{\hat{S}}^{(s)} + \sqrt{\tau_i} \cdot \mathbf{S}^{(s)},
\end{equation}
where $\tau_i = \hat{t}_i - t^q_i$ denotes the fractional offset from the preceding temporal anchor. And for latent space $H=32, W=32, C=4$ and a model video of $V=16$ frames.

This strategy produces a structured noise tensor that is both \emph{viewpoint-consistent} and \emph{temporally smooth}, thereby enhancing the stability of optimization and improving the realism of generated 4D content.
This approach maintains coherence in the dynamic effects while preserving spatial correspondence, allowing the generated motion to remain consistent regardless of the viewing angle.

\subsection{Attention-masked SDS}
\label{sec:attention_SDS}
In our approach, since the 3D initial object already exists, we need to focus the loss specifically on the regions undergoing growth.
In contrast, the standard implementation of SDS loss computes it over the entire object’s latent representation. By leveraging attention maps, we can guide the learning process toward the most relevant regions, ensuring that optimization is focused on the object. 
This approach ultimately enhances subject preservation while capturing more dynamic changes.
For example, in Fig.\ref{fig:pipeline}, the attention masks highlight higher values in the branches, where growth is occurring, while the pot has lower values since it remains unchanged.
In our approach, we found that the first cross-attention mask between the input object rendering and the NeRF renders provides the most accurate masking, best highlighting the regions where growth occurs.
The masked SDS loss is the pointwise product:
\begin{equation}
    \mathcal{L}_{masked-SDS} = M  \mathcal{L}_{I2V} , 
    \label{masked-SDS}
\end{equation}
where $M$ is the attention mask.

\subsection{Modeling time}
\label{sec:time_sampling}
Video generative models typically operate on $V=16$ frame sequences, but we aim for a 4D representation that can generate videos at any frame count, ensuring smooth and continuous dynamics without fixed frame limits. To achieve this, we currently sample a video from the NeRF by selecting a starting time $t_0=\mathcal{U}[0, 1/V]$ and uniformly sampling more $V-1$ frames from the range, $[t_0, 1]$, allowing continuous sampling \cite{mav3d, 4dfy}. 
However, this approach is suboptimal for image-to-video models, as it forces the static object to remain across all time steps, limiting dynamics. Additionally, the initial frame $t=0$, this frame is rarely selected.

In \texttt{\ourmethod{}} we propose a new time sampling strategy: we evenly select 16 frame times within the range $\left[ 0, 1 \right]$, with the first frame fixed at $t_0 = 0$ and later frame times adjusted with small noise $\hat{t}_i = \nicefrac{i}{V} + \epsilon_i$. 
Our time sampling strategy ensures uniform sampling across the entire time range while maintaining the input object condition requirements.

\section{Experiments}
\label{sec:experiments}

\subsection{Dataset}
We used two public datasets of 3D objects in our experiments. The first is the Google Scanned Objects (GSO) dataset \citep{google-scanned-objects}, which consists of high-quality 3D scans of everyday items. The second is the Objaverse dataset \citep{objaverse}, which contains a large-scale collection of diverse 3D assets gathered from various sources.
We selected objects from the GSO dataset and from objaverse, focusing on those that could support interesting growth dynamics and motion. For this purpose, we queried ChatGPT for objects and corresponding prompts that elicit significant non-rigid deformations. This resulted in a selection of 20 objects from GSO and 10 from Objaverse.

\subsection{Metrics}
\label{sec:metrics}
To evaluate our approach, we assess three main qualities: (1) preservation of the input object's identity, (2) natural appearance of the generated 4D content, and (3) alignment with the text prompt.
We use four evaluation metrics from Vbench \citep{Vbench}.

\textbf{(1) Motion Smoothness.} We assess whether the motion in the generated video is smooth. To do so, we leverage motion priors from the video frame interpolation model \citep{amt} ("smoothness"), as suggested in VBench \citep{Vbench}. 
\textbf{(2) Dynamic Degree.} Since static objects can also exhibit high motion smoothness, we introduce an additional metric to evaluate the presence of dynamic content in the video. Specifically, we quantify the amount of movement by computing optical flow between frames using RAFT \citep{raft}, following the protocol in VBench \citep{Vbench}.
\textbf{(3) Agreement with prompt.} To measure consistency with the prompt, we followed Vbench \citep{Vbench} and measured the similarity between the video frame features and the textual description features with ViCLIP \citep{viclip} ("style").
\textbf{(4) Agreement with input object.} Ensuring visual consistency between the input 3D object and the generated 4D data. To evaluate this, we used LPIPS \citep{lpips} to measure the perceptual similarity between the input object renders and the generated frames, assessing the consistency of visual appearance over time.

We compute all metrics across four viewpoints with azimuth angles ${0^\circ, 90^\circ, 180^\circ, 270^\circ}$ and fixed elevation $0^\circ$, averaging scores per object across views. Final results are reported as the mean and standard error across all objects.

\begin{figure}[t]
    \centering
    \includegraphics[width=\textwidth]{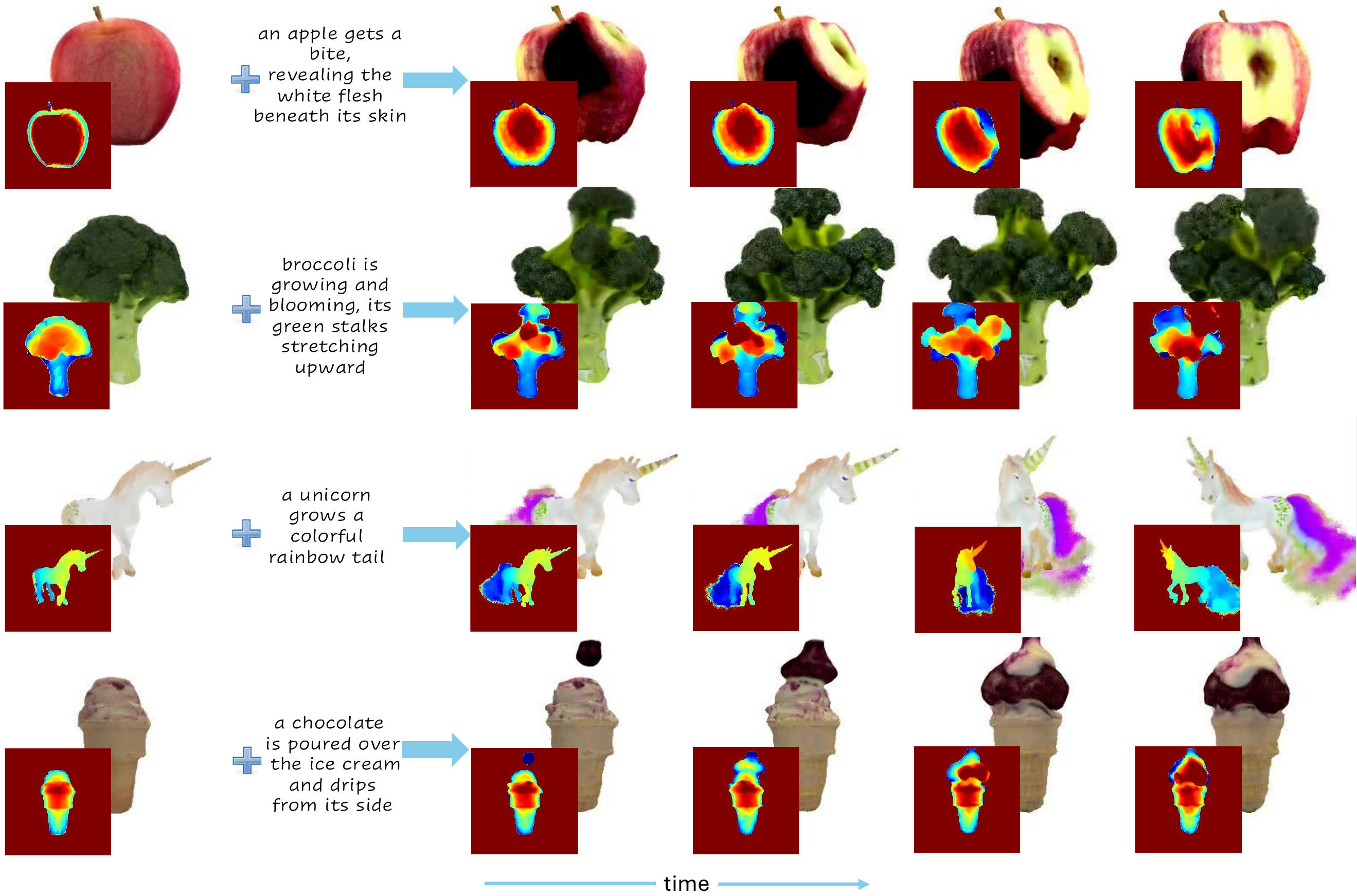}
    \caption{\textbf{\ourmethod{}} brings various objects to life. On the left, we display the input object along with a textual prompt describing the desired action. On the right, we present four frames from the generated object, viewed from the front. Each 3D frame is split into an RGB image and its corresponding depth map, shown in the top right corner.} 
    \label{fig:qualitative}
\end{figure}

\begin{table}[t]
    \centering

    \begin{subtable}[t]{\textwidth}
        \centering
        \resizebox{\textwidth}{!}{%
        \begin{tabular}{lcccc}
            \toprule
            \multicolumn{5}{c}{\textbf{Google Scanned Objects}} \\
            \midrule
            \textbf{Method} & \textbf{Temporal style} $\uparrow$ & \textbf{Motion smoothness} $\uparrow$ & \textbf{Dynamic degree} $\uparrow$ & \textbf{Identity Preservation} $\downarrow$ \\
            \midrule
            Ours & \textbf{17.8 $\pm$ 1.3} & 99.1 $\pm$ 0.07 & \textbf{50.5 $\pm$ 2.1} & 18.9 $\pm$ 2.0 \\
            Animate3D & 13.7 $\pm$ 1.0 & \textbf{99.2 $\pm$ 0.008} & 21.5 $\pm$ 4.8 & \textbf{17.6 $\pm$ 2.1} \\
            \bottomrule
        \end{tabular}
        }
    \end{subtable}

    \vspace{0.1em}

    \begin{subtable}[t]{\textwidth}
        \centering
        \resizebox{\textwidth}{!}{%
        \begin{tabular}{lcccc}
            \toprule
            \multicolumn{5}{c}{\textbf{Objaverse}} \\
            \midrule
            \textbf{Method} & \textbf{Temporal style} $\uparrow$ & \textbf{Motion smoothness} $\uparrow$ & \textbf{Dynamic degree} $\uparrow$ & \textbf{Identity Preservation} $\downarrow$ \\
            \midrule
            Ours & \textbf{17.2 $\pm$ 1.4} & 99.1 $\pm$ 0.12 & \textbf{51.7 $\pm$ 7.5} & 19.4 $\pm$ 3.2 \\
            Animate3D & 16.3 $\pm$ 1.5 & \textbf{99.3 $\pm$ 0.01} & 27.7 $\pm$ 1.2 & \textbf{15.8 $\pm$ 3.0} \\
            \bottomrule
        \end{tabular}
        }
    \end{subtable}

    \caption{
    Comparison between \ourmethod{} and Animate3D. 
    Metrics are explained in \ref{sec:metrics}.
    Our \ourmethod{} excels in dynamic degree and agreement with the prompt, while Animate3D better preserves object identity, as it does not attempt to evolve the objects—such as the melting of ice in the cream shown in Figure~\ref{fig:teaser}.
    }
    \label{text_vs_image}
\end{table}

\subsection{Compared methods}
\label{seq:compared_methods}
Two previous studies have animated 3D objects using multi-view diffusion models, \textit{Animate3D}~\cite{animate3d} and \textit{Diffusion4D}~\cite{diffusion4d}. Unfortunately, Diffusion4D did not released their code, and we can only present comparisons with Animate3D.

\subsection{Implementation details}
\label{sec: implementation}
We implement Image-to-Video SDS using the ThreeStudio framework \citep{threestudio}. Our implementation builds upon the text-to-4D capabilities of ThreeStudio \citep{4dfy}, replacing its viewpoint sampling protocol with the method proposed by \cite{sdscomplete}.

\textbf{Networks and rendering:} We used a hash encoding-based neural representation, following the implementation in \citep{4dfy}. For image-to-video model, we used DynamiCrafter \citep{dynamicrafter}, which generates videos at a resolution of 256x256. The input 3D object was rendered using PyTorch3D \citep{pytorch3d}, matching DynamiCrafter resolution with a rendering size of 256x256. The number of frames is $V=16$.

\textbf{Running Time:} Our NeRF representation conversion was performed over 5000 iterations with uniform viewpoint sampling, taking approximately 10 minutes on an NVIDIA H100 GPU. The second phases was run for 20,000 steps and took $\sim 240$ minutes.

\section{Results}
\label{Results}
We first provide qualitative examples of \ourmethod{}, then a quantitative and qualitative comparison of \ourmethod{} with the baselines methods. Finally, quantitative and qualitative results of an ablation study, and the effect of different prompts.

\textbf{Qualitative results:}
Figure \ref{fig:qualitative} shows four examples of 4D generations (right) from a 3D object (left). 

\textbf{Quantitative comparison with baselines: } 
Table~(\ref{text_vs_image}) compares \ourmethod{} with the three baselines. \ourmethod{} achieves far better agreement with the input object (identity preservation) in both LPIPS. It also generates a slightly more smooth and natural-looking 4D than other baselines. 
Agreement with the prompt is lower, presumably because the content adheres to the input object, which may deviate from the canonical representation of the corresponding text term. In other words, the text prompt may push an object to have other appearance than the given input object.   

\begin{table}[t]
    \centering

    \begin{subtable}[t]{\textwidth}
        \centering
        \resizebox{\textwidth}{!}{%
        \begin{tabular}{lcccc}
            \toprule
            \multicolumn{5}{c}{\textbf{Google Scanned Objects (20 objects)}} \\
            \midrule
            \textbf{Method} & \textbf{Temporal style} $\uparrow$ & \textbf{Motion smoothness} $\uparrow$ & \textbf{Dynamic degree} $\uparrow$ & \textbf{Identity Preservation} $\downarrow$ \\
            \midrule
            Ours (full) & \textbf{17.8 $\pm$ 1.3} & \textbf{99.1 $\pm$ 0.07} & \textbf{50.5 $\pm$ 2.1} & \textbf{18.9 $\pm$ 2.0} \\
            w.o. view consistency & 16.5 $\pm$ 1.0 & 99.1 $\pm$ 0.06 & 41.4 $\pm$ 2.0 & 18.9 $\pm$ 1.8 \\
            w.o. masked SDS & 17.0 $\pm$ 1.1 & 99.1 $\pm$ 0.06 & 46.6 $\pm$ 3.0 & 19.8 $\pm$ 3.1 \\
            \bottomrule
        \end{tabular}
        }
    \end{subtable}

    \vspace{0.1em}

    \begin{subtable}[t]{\textwidth}
        \centering
        \resizebox{\textwidth}{!}{%
        \begin{tabular}{lcccc}
            \toprule
            \multicolumn{5}{c}{\textbf{Objaverse (10 objects)}} \\
            \midrule
            \textbf{Method} & \textbf{Temporal style} $\uparrow$ & \textbf{Motion smoothness} $\uparrow$ & \textbf{Dynamic degree} $\uparrow$ & \textbf{Subject consistency} $\downarrow$ \\
            \midrule
            Ours (full) & 17.2 $\pm$ 1.4 & \textbf{99.2 $\pm$ 0.08} & \textbf{51.7 $\pm$ 7.5} & 19.4 $\pm$ 3.2 \\
            w.o. view consistency & \textbf{17.5 $\pm$ 1.3} & 99.1 $\pm$ 0.1 &  45.3 $\pm$ 3.9 & 18.5 $\pm$ 3.3 \\
            w.o. masked SDS & 13.7 $\pm$ 1.3 & 99.1 $\pm$ 0.1 & 48.7 $\pm$ 4.4 & \textbf{17.7 $\pm$ 3.2} \\
            \bottomrule
        \end{tabular}
        }
    \end{subtable}

    \caption{\textbf{Ablation study.} Evaluating the contribution of various components of our method.}   
    \label{tab:ablation}
\end{table}

\begin{figure}[t]
    Input object \hspace{100pt}
    \ourmethod{} (ours) \hspace{90pt}
    Animate3D  \hspace{50pt} \\
    \centering
    \includegraphics[width=0.96\linewidth]{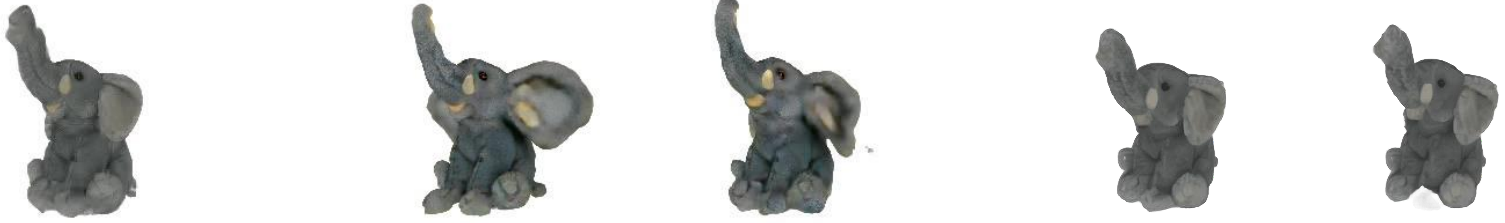}
    \includegraphics[width=0.96\linewidth]{Figures/comparison_basline-2-crop.pdf}
    \caption{Qualitative comparison. A render of the input object is shown on the left, alongside renders from \ourmethod{} (middle) and Animate3D (right). In this example, our method generates a 4D object that is better aligned with the prompt "\textit{an elephant grows its ears as long as wings to fly},".} 
    \label{fig:comparisons}
\end{figure}

\begin{figure}[h]
    \centering
    \includegraphics[width=0.98\linewidth]{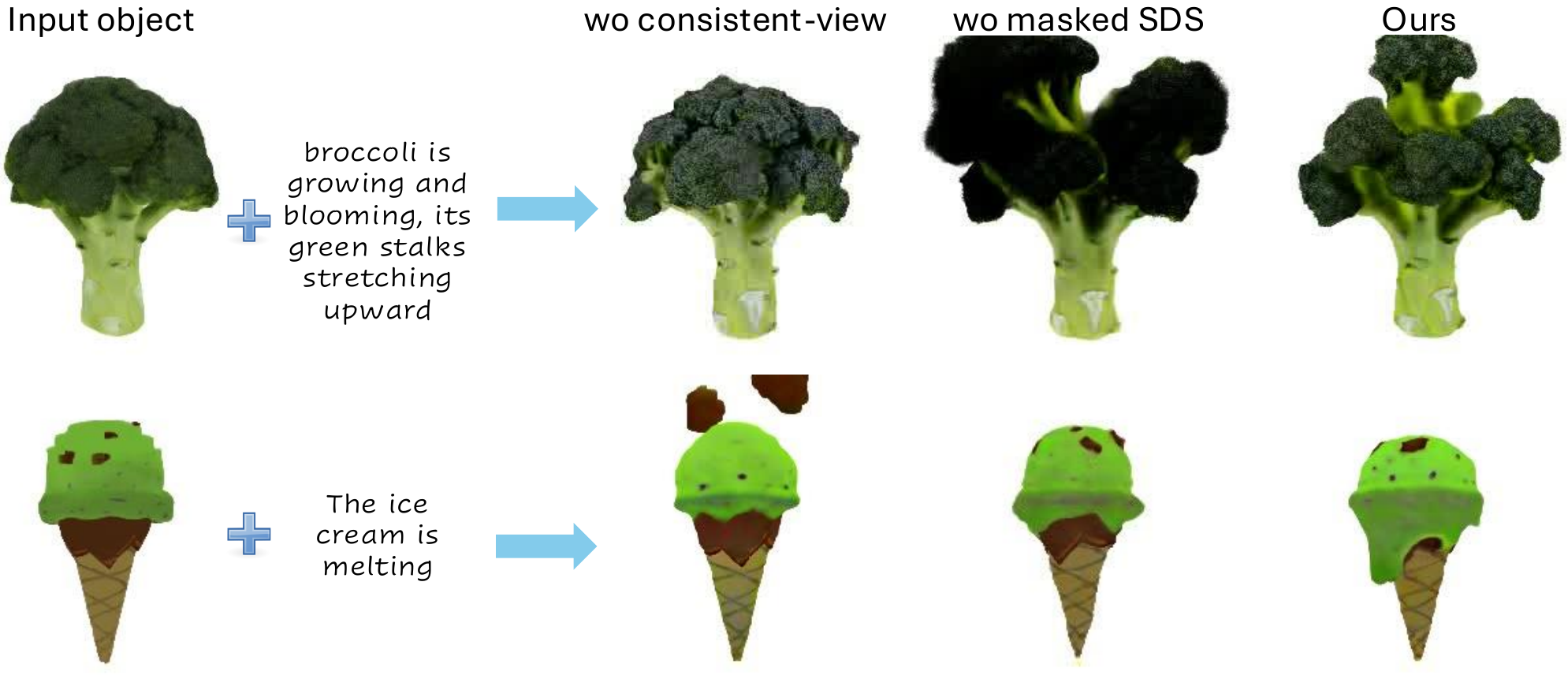}
    \caption{Qualitative ablation results demonstrate the contribution of each part of our method. Without our view-consistency noising the broccoli does not ``bloom". Without our attention-masked SDS, the plant is less rich in details.} 
    \label{fig:ablation}
\end{figure}

\textbf{Qualitative comparisons:}
Figure~\ref{fig:comparisons} presents a qualitative comparison between \ourmethod{} and Animate3D. While \ourmethod{} aligns closely with the prompt and generates high dynamic motion, Animate3D fails to follow the prompt and produces a 4D output with limited dynamics.

\noindent\textbf{Ablation analysis:}
We conducted an ablation study to evaluate the contributions of each component of \ourmethod{}. Table \ref{tab:ablation} provides the quantitative results. Without view-consist-noise, the dynamic degree is strongly reduced, because the model tends to average across inconsistent videos. The dynamic degree is also hurt without attention-masked SDS
Altogether, \ourmethod{} achieves a balanced trade-off between preserving the input object's identity, fulfilling the prompt, and maintaining a high dynamic degree.
Figure~\ref{fig:ablation} illustrates the ablation effect using two qualitative examples. 

\subsection{Noise consistency and video consistency}

\begin{wrapfigure}{r}{0.5\textwidth}
    \vspace{-30pt}
    \centering
    \includegraphics[width=0.5\textwidth]{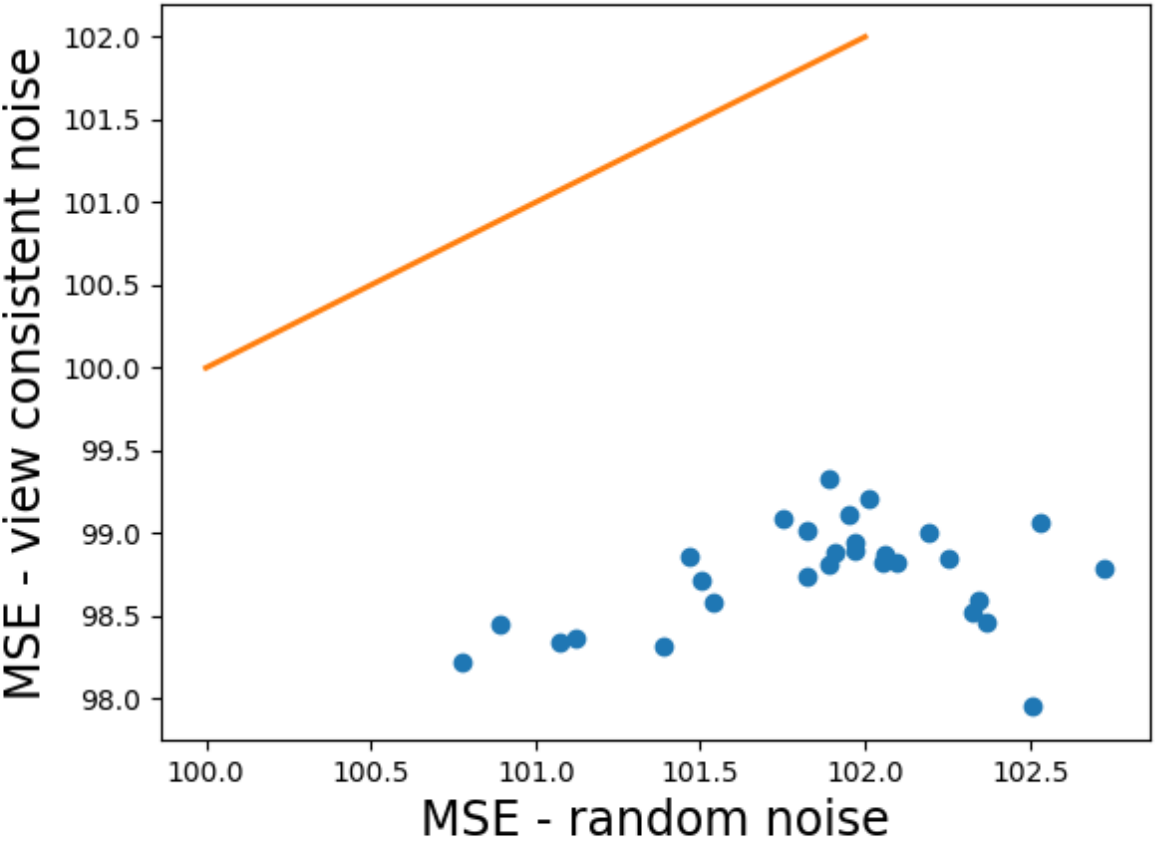}
    \caption{ MSE across video pairs from near viewpoints, using view-consistent noise (y-axis) and random noise (x-axis). Yellow line represents the equality (y=x). Each dot denotes one object tested. View-consistent noise results in a lower mean MSE across all objects. }
    \label{fig:comparison}
\end{wrapfigure}

To gain more insight into the effect of view-consistency noising in video generation across different viewing angles, we render images of the objects from angles ${0^{\circ}, 5^{\circ}, \dots, 355^{\circ}}$, each with corresponding view-consistent noise. We then generate a video for each angle and compute the MSE between adjacent video pairs, i.e., ${(0^{\circ}, 5^{\circ}), (5^{\circ}, 10^{\circ}), \dots, (350^{\circ}, 355^{\circ})}$. We then repeated this process, but this time using random noise. 

Figure~\ref{fig:comparison} presents two video MSE values: from the view-consistent noise (y-axis) and random noise (x-axis). The yellow line is the identity function $y=x$. Each dot denotes the average MSE for one object. The view-consistent noise achieves lower MSE across all objects, demonstrating that it encourages the text-to-video model to generate more consistent appearance and movement across camera viewpoints.

\section{Conclusion}
We present \ourmethod{}, a novel method for animating 3D objects into dynamic 4D scenes from textual motion prompts. It uses Image-to-Video diffusion models, ensuring object consistency via rendered image conditioning and a tailored SDS loss. To boost motion realism, we introduce a view-consistent noise and an attention-guided masked SDS loss. \ourmethod{} achieves better, prompt alignment, and visual fidelity, offering an effective solution for controlled 4D content creation.

\textbf{Limitations:} \ourmethod{} builds on given video-generation models, and therefore inherits their underlying limitations such as limb confusion and missing object parts. Also, our current implementation has a large memory footprint, which may not work with new video generation models.

{
    \small
    \bibliographystyle{plainnat}
    \bibliography{Arxiv}

\begin{thebibliography}{47}
\providecommand{\natexlab}[1]{#1}
\providecommand{\url}[1]{\texttt{#1}}
\expandafter\ifx\csname urlstyle\endcsname\relax
  \providecommand{\doi}[1]{doi: #1}\else
  \providecommand{\doi}{doi: \begingroup \urlstyle{rm}\Url}\fi

\bibitem[Bahmani et~al.(2024{\natexlab{a}})Bahmani, Liu, Yifan, Skorokhodov, Rong, Liu, Liu, Park, Tulyakov, Wetzstein, et~al.]{tc4d}
Sherwin Bahmani, Xian Liu, Wang Yifan, Ivan Skorokhodov, Victor Rong, Ziwei Liu, Xihui Liu, Jeong~Joon Park, Sergey Tulyakov, Gordon Wetzstein, et~al.
\newblock Tc4d: Trajectory-conditioned text-to-4d generation.
\newblock In \emph{European Conference on Computer Vision}, pages 53--72. Springer, 2024{\natexlab{a}}.

\bibitem[Bahmani et~al.(2024{\natexlab{b}})Bahmani, Skorokhodov, Rong, Wetzstein, Guibas, Wonka, Tulyakov, Park, Tagliasacchi, and Lindell]{4dfy}
Sherwin Bahmani, Ivan Skorokhodov, Victor Rong, Gordon Wetzstein, Leonidas Guibas, Peter Wonka, Sergey Tulyakov, Jeong~Joon Park, Andrea Tagliasacchi, and David~B Lindell.
\newblock 4d-fy: Text-to-4d generation using hybrid score distillation sampling.
\newblock In \emph{Proceedings of the IEEE/CVF Conference on Computer Vision and Pattern Recognition}, pages 7996--8006, 2024{\natexlab{b}}.

\bibitem[Blattmann et~al.(2023)Blattmann, Dockhorn, Kulal, Mendelevitch, Kilian, Lorenz, Levi, English, Voleti, Letts, et~al.]{stableVideo}
Andreas Blattmann, Tim Dockhorn, Sumith Kulal, Daniel Mendelevitch, Maciej Kilian, Dominik Lorenz, Yam Levi, Zion English, Vikram Voleti, Adam Letts, et~al.
\newblock Stable video diffusion: Scaling latent video diffusion models to large datasets.
\newblock \emph{arXiv preprint arXiv:2311.15127}, 2023.

\bibitem[Chu et~al.(2024)Chu, Ke, and Fragkiadaki]{dreamscene4d}
Wen-Hsuan Chu, Lei Ke, and Katerina Fragkiadaki.
\newblock Dreamscene4d: Dynamic multi-object scene generation from monocular videos.
\newblock \emph{arXiv preprint arXiv:2405.02280}, 2024.

\bibitem[Deitke et~al.(2023)Deitke, Schwenk, Salvador, Weihs, Michel, VanderBilt, Schmidt, Ehsani, Kembhavi, and Farhadi]{objaverse}
Matt Deitke, Dustin Schwenk, Jordi Salvador, Luca Weihs, Oscar Michel, Eli VanderBilt, Ludwig Schmidt, Kiana Ehsani, Aniruddha Kembhavi, and Ali Farhadi.
\newblock Objaverse: A universe of annotated 3d objects.
\newblock In \emph{Proceedings of the IEEE/CVF Conference on Computer Vision and Pattern Recognition}, pages 13142--13153, 2023.

\bibitem[Deng et~al.(2025)Deng, Xiong, Feng, Wang, and Liu]{stp4d}
Yunze Deng, Haijun Xiong, Bin Feng, Xinggang Wang, and Wenyu Liu.
\newblock Stp4d: Spatio-temporal-prompt consistent modeling for text-to-4d gaussian splatting.
\newblock \emph{arXiv preprint arXiv:2504.18318}, 2025.

\bibitem[Downs et~al.(2022)Downs, Francis, Koenig, Kinman, Hickman, Reymann, McHugh, and Vanhoucke]{google-scanned-objects}
Laura Downs, Anthony Francis, Nate Koenig, Brandon Kinman, Ryan Hickman, Krista Reymann, Thomas~B McHugh, and Vincent Vanhoucke.
\newblock Google scanned objects: A high-quality dataset of 3d scanned household items.
\newblock In \emph{2022 International Conference on Robotics and Automation (ICRA)}, pages 2553--2560. IEEE, 2022.

\bibitem[Gao et~al.(2024)Gao, Xu, Cao, Mildenhall, Ma, Chen, Tang, and Neumann]{gaussianflow}
Quankai Gao, Qiangeng Xu, Zhe Cao, Ben Mildenhall, Wenchao Ma, Le~Chen, Danhang Tang, and Ulrich Neumann.
\newblock Gaussianflow: Splatting gaussian dynamics for 4d content creation.
\newblock \emph{arXiv preprint arXiv:2403.12365}, 2024.

\bibitem[Guo et~al.(2023)Guo, Liu, Shao, Laforte, Voleti, Luo, Chen, Zou, Wang, Cao, and Zhang]{threestudio}
Yuan-Chen Guo, Ying-Tian Liu, Ruizhi Shao, Christian Laforte, Vikram Voleti, Guan Luo, Chia-Hao Chen, Zi-Xin Zou, Chen Wang, Yan-Pei Cao, and Song-Hai Zhang.
\newblock threestudio: A unified framework for 3d content generation.
\newblock \url{https://github.com/threestudio-project/threestudio}, 2023.

\bibitem[HaCohen et~al.(2024)HaCohen, Chiprut, Brazowski, Shalem, Moshe, Richardson, Levin, Shiran, Zabari, Gordon, et~al.]{ltxv}
Yoav HaCohen, Nisan Chiprut, Benny Brazowski, Daniel Shalem, Dudu Moshe, Eitan Richardson, Eran Levin, Guy Shiran, Nir Zabari, Ori Gordon, et~al.
\newblock Ltx-video: Realtime video latent diffusion.
\newblock \emph{arXiv preprint arXiv:2501.00103}, 2024.

\bibitem[Ho et~al.(2022)Ho, Chan, Saharia, Whang, Gao, Gritsenko, Kingma, Poole, Norouzi, Fleet, et~al.]{imagenvideo}
Jonathan Ho, William Chan, Chitwan Saharia, Jay Whang, Ruiqi Gao, Alexey Gritsenko, Diederik~P Kingma, Ben Poole, Mohammad Norouzi, David~J Fleet, et~al.
\newblock Imagen video: High definition video generation with diffusion models.
\newblock \emph{arXiv preprint arXiv:2210.02303}, 2022.

\bibitem[Huang et~al.(2024)Huang, He, Yu, Zhang, Si, Jiang, Zhang, Wu, Jin, Chanpaisit, et~al.]{Vbench}
Ziqi Huang, Yinan He, Jiashuo Yu, Fan Zhang, Chenyang Si, Yuming Jiang, Yuanhan Zhang, Tianxing Wu, Qingyang Jin, Nattapol Chanpaisit, et~al.
\newblock Vbench: Comprehensive benchmark suite for video generative models.
\newblock In \emph{Proceedings of the IEEE/CVF Conference on Computer Vision and Pattern Recognition}, pages 21807--21818, 2024.

\bibitem[Jiang et~al.(2024)Jiang, Yu, Cao, Wang, Hu, and Gao]{animate3d}
Yanqin Jiang, Chaohui Yu, Chenjie Cao, Fan Wang, Weiming Hu, and Jin Gao.
\newblock Animate3d: Animating any 3d model with multi-view video diffusion.
\newblock \emph{arXiv preprint arXiv:2407.11398}, 2024.

\bibitem[Kasten et~al.(2024)Kasten, Rahamim, and Chechik]{sdscomplete}
Yoni Kasten, Ohad Rahamim, and Gal Chechik.
\newblock Point cloud completion with pretrained text-to-image diffusion models.
\newblock \emph{Advances in Neural Information Processing Systems}, 36, 2024.

\bibitem[Li et~al.(2024{\natexlab{a}})Li, Pan, Yang, Xu, Zhou, Zhang, Li, Kadambi, Wang, Tu, et~al.]{4k4dgen}
Renjie Li, Panwang Pan, Bangbang Yang, Dejia Xu, Shijie Zhou, Xuanyang Zhang, Zeming Li, Achuta Kadambi, Zhangyang Wang, Zhengzhong Tu, et~al.
\newblock 4k4dgen: Panoramic 4d generation at 4k resolution.
\newblock \emph{arXiv preprint arXiv:2406.13527}, 2024{\natexlab{a}}.

\bibitem[Li et~al.(2023)Li, Zhu, Han, Hou, Guo, and Cheng]{amt}
Zhen Li, Zuo-Liang Zhu, Ling-Hao Han, Qibin Hou, Chun-Le Guo, and Ming-Ming Cheng.
\newblock Amt: All-pairs multi-field transforms for efficient frame interpolation.
\newblock In \emph{Proceedings of the IEEE/CVF Conference on Computer Vision and Pattern Recognition}, pages 9801--9810, 2023.

\bibitem[Li et~al.(2024{\natexlab{b}})Li, Chen, and Liu]{dreammesh4d}
Zhiqi Li, Yiming Chen, and Peidong Liu.
\newblock Dreammesh4d: Video-to-4d generation with sparse-controlled gaussian-mesh hybrid representation.
\newblock \emph{Advances in Neural Information Processing Systems}, 37:\penalty0 21377--21400, 2024{\natexlab{b}}.

\bibitem[Liang et~al.(2024)Liang, Yin, Xu, Liang, Wang, Plataniotis, Zhao, and Wei]{diffusion4d}
Hanwen Liang, Yuyang Yin, Dejia Xu, Hanxue Liang, Zhangyang Wang, Konstantinos~N Plataniotis, Yao Zhao, and Yunchao Wei.
\newblock Diffusion4d: Fast spatial-temporal consistent 4d generation via video diffusion models.
\newblock \emph{arXiv preprint arXiv:2405.16645}, 2024.

\bibitem[Ling et~al.(2024)Ling, Kim, Torralba, Fidler, and Kreis]{aligngaussians}
Huan Ling, Seung~Wook Kim, Antonio Torralba, Sanja Fidler, and Karsten Kreis.
\newblock Align your gaussians: Text-to-4d with dynamic 3d gaussians and composed diffusion models.
\newblock In \emph{Proceedings of the IEEE/CVF Conference on Computer Vision and Pattern Recognition}, pages 8576--8588, 2024.

\bibitem[Liu and Vahdat(2025)]{equivdm}
Chao Liu and Arash Vahdat.
\newblock Equivdm: Equivariant video diffusion models with temporally consistent noise.
\newblock \emph{arXiv preprint arXiv:2504.09789}, 2025.

\bibitem[Miao et~al.(2024)Miao, Quan, Li, and Luo]{pla4d}
Qiaowei Miao, JinSheng Quan, Kehan Li, and Yawei Luo.
\newblock Pla4d: Pixel-level alignments for text-to-4d gaussian splatting.
\newblock \emph{arXiv preprint arXiv:2405.19957}, 2024.

\bibitem[Nag et~al.(2025)Nag, Cohen-Or, Zhang, and Mahdavi-Amiri]{in24d}
Sauradip Nag, Daniel Cohen-Or, Hao Zhang, and Ali Mahdavi-Amiri.
\newblock In-2-4d: Inbetweening from two single-view images to 4d generation.
\newblock \emph{arXiv preprint arXiv:2504.08366}, 2025.

\bibitem[Park et~al.(2025)Park, Kwon, and Ye]{zero4d}
Jangho Park, Taesung Kwon, and Jong~Chul Ye.
\newblock Zero4d: Training-free 4d video generation from single video using off-the-shelf video diffusion model.
\newblock \emph{arXiv preprint arXiv:2503.22622}, 2025.

\bibitem[Poole et~al.(2022)Poole, Jain, Barron, and Mildenhall]{dreamfusion}
Ben Poole, Ajay Jain, Jonathan~T Barron, and Ben Mildenhall.
\newblock Dreamfusion: Text-to-3d using 2d diffusion.
\newblock \emph{arXiv preprint arXiv:2209.14988}, 2022.

\bibitem[Radford et~al.(2021)Radford, Kim, Hallacy, Ramesh, Goh, Agarwal, Sastry, Askell, Mishkin, Clark, et~al.]{viclip}
Alec Radford, Jong~Wook Kim, Chris Hallacy, Aditya Ramesh, Gabriel Goh, Sandhini Agarwal, Girish Sastry, Amanda Askell, Pamela Mishkin, Jack Clark, et~al.
\newblock Learning transferable visual models from natural language supervision.
\newblock In \emph{International conference on machine learning}, pages 8748--8763. PMLR, 2021.

\bibitem[Ravi et~al.(2020)Ravi, Reizenstein, Novotny, Gordon, Lo, Johnson, and Gkioxari]{pytorch3d}
Nikhila Ravi, Jeremy Reizenstein, David Novotny, Taylor Gordon, Wan-Yen Lo, Justin Johnson, and Georgia Gkioxari.
\newblock Accelerating 3d deep learning with pytorch3d.
\newblock \emph{arXiv:2007.08501}, 2020.

\bibitem[Ren et~al.(2023)Ren, Pan, Tang, Zhang, Cao, Zeng, and Liu]{dreamgaussian4d}
Jiawei Ren, Liang Pan, Jiaxiang Tang, Chi Zhang, Ang Cao, Gang Zeng, and Ziwei Liu.
\newblock Dreamgaussian4d: Generative 4d gaussian splatting.
\newblock \emph{arXiv preprint arXiv:2312.17142}, 2023.

\bibitem[Ren et~al.(2025)Ren, Xie, Mirzaei, Kreis, Liu, Torralba, Fidler, Kim, Ling, et~al.]{l4gm}
Jiawei Ren, Cheng Xie, Ashkan Mirzaei, Karsten Kreis, Ziwei Liu, Antonio Torralba, Sanja Fidler, Seung~Wook Kim, Huan Ling, et~al.
\newblock L4gm: Large 4d gaussian reconstruction model.
\newblock \emph{Advances in Neural Information Processing Systems}, 37:\penalty0 56828--56858, 2025.

\bibitem[Sang et~al.(2025)Sang, Canfes, Cao, Marin, Bernard, and Cremers]{twosquared}
Lu~Sang, Zehranaz Canfes, Dongliang Cao, Riccardo Marin, Florian Bernard, and Daniel Cremers.
\newblock Twosquared: 4d generation from 2d image pairs.
\newblock \emph{arXiv preprint arXiv:2504.12825}, 2025.

\bibitem[Singer et~al.(2023)Singer, Sheynin, Polyak, Ashual, Makarov, Kokkinos, Goyal, Vedaldi, Parikh, Johnson, et~al.]{mav3d}
Uriel Singer, Shelly Sheynin, Adam Polyak, Oron Ashual, Iurii Makarov, Filippos Kokkinos, Naman Goyal, Andrea Vedaldi, Devi Parikh, Justin Johnson, et~al.
\newblock Text-to-4d dynamic scene generation.
\newblock \emph{arXiv preprint arXiv:2301.11280}, 2023.

\bibitem[Sun et~al.(2024)Sun, Chen, Liu, Chen, Duan, Zhang, and Wang]{dimensionx}
Wenqiang Sun, Shuo Chen, Fangfu Liu, Zilong Chen, Yueqi Duan, Jun Zhang, and Yikai Wang.
\newblock Dimensionx: Create any 3d and 4d scenes from a single image with controllable video diffusion.
\newblock \emph{arXiv preprint arXiv:2411.04928}, 2024.

\bibitem[Teed and Deng(2020)]{raft}
Zachary Teed and Jia Deng.
\newblock Raft: Recurrent all-pairs field transforms for optical flow.
\newblock In \emph{Computer Vision--ECCV 2020: 16th European Conference, Glasgow, UK, August 23--28, 2020, Proceedings, Part II 16}, pages 402--419. Springer, 2020.

\bibitem[Wu et~al.(2024)Wu, Yu, Jiang, Cao, Wang, and Bai]{sc4d}
Zijie Wu, Chaohui Yu, Yanqin Jiang, Chenjie Cao, Fan Wang, and Xiang Bai.
\newblock Sc4d: Sparse-controlled video-to-4d generation and motion transfer.
\newblock In \emph{European Conference on Computer Vision}, pages 361--379. Springer, 2024.

\bibitem[Xie et~al.(2024)Xie, Yao, Voleti, Jiang, and Jampani]{sv4d}
Yiming Xie, Chun-Han Yao, Vikram Voleti, Huaizu Jiang, and Varun Jampani.
\newblock Sv4d: Dynamic 3d content generation with multi-frame and multi-view consistency.
\newblock \emph{arXiv preprint arXiv:2407.17470}, 2024.

\bibitem[Xing et~al.(2023)Xing, Xia, Zhang, Chen, Wang, Wong, and Shan]{dynamicrafter}
Jinbo Xing, Menghan Xia, Yong Zhang, Haoxin Chen, Xintao Wang, Tien-Tsin Wong, and Ying Shan.
\newblock Dynamicrafter: Animating open-domain images with video diffusion priors.
\newblock \emph{arXiv preprint arXiv:2310.12190}, 2023.

\bibitem[Xu et~al.(2024)Xu, Liang, Bhatt, Hu, Liang, Plataniotis, and Wang]{comp4d}
Dejia Xu, Hanwen Liang, Neel~P Bhatt, Hezhen Hu, Hanxue Liang, Konstantinos~N Plataniotis, and Zhangyang Wang.
\newblock Comp4d: Llm-guided compositional 4d scene generation.
\newblock \emph{arXiv preprint arXiv:2403.16993}, 2024.

\bibitem[Yan et~al.(2025)Yan, Chen, and Wang]{CFD}
Runjie Yan, Yinbo Chen, and Xiaolong Wang.
\newblock Consistent flow distillation for text-to-3d generation.
\newblock \emph{arXiv preprint arXiv:2501.05445}, 2025.

\bibitem[Yang et~al.(2025)Yang, Liu, Zhu, Liu, Ma, Nong, and Liang]{yang2025not}
Liying Yang, Chen Liu, Zhenwei Zhu, Ajian Liu, Hui Ma, Jian Nong, and Yanyan Liang.
\newblock Not all frame features are equal: Video-to-4d generation via decoupling dynamic-static features.
\newblock \emph{arXiv preprint arXiv:2502.08377}, 2025.

\bibitem[Yao et~al.(2025)Yao, Xie, Voleti, Jiang, and Jampani]{sv4d2}
Chun-Han Yao, Yiming Xie, Vikram Voleti, Huaizu Jiang, and Varun Jampani.
\newblock Sv4d 2.0: Enhancing spatio-temporal consistency in multi-view video diffusion for high-quality 4d generation.
\newblock \emph{arXiv preprint arXiv:2503.16396}, 2025.

\bibitem[Yin et~al.(2023)Yin, Xu, Wang, Zhao, and Wei]{4dgen}
Yuyang Yin, Dejia Xu, Zhangyang Wang, Yao Zhao, and Yunchao Wei.
\newblock 4dgen: Grounded 4d content generation with spatial-temporal consistency.
\newblock \emph{arXiv preprint arXiv:2312.17225}, 2023.

\bibitem[Yu et~al.(2024)Yu, Wang, Zhuang, Menapace, Siarohin, Cao, Jeni, Tulyakov, and Lee]{4real}
Heng Yu, Chaoyang Wang, Peiye Zhuang, Willi Menapace, Aliaksandr Siarohin, Junli Cao, L{\'a}szl{\'o} Jeni, Sergey Tulyakov, and Hsin-Ying Lee.
\newblock 4real: Towards photorealistic 4d scene generation via video diffusion models.
\newblock \emph{Advances in Neural Information Processing Systems}, 37:\penalty0 45256--45280, 2024.

\bibitem[Yuan et~al.(2024)Yuan, Kobbelt, Liu, Zhang, Wan, Lai, and Gao]{4dynamic}
Yu-Jie Yuan, Leif Kobbelt, Jiwen Liu, Yuan Zhang, Pengfei Wan, Yu-Kun Lai, and Lin Gao.
\newblock 4dynamic: Text-to-4d generation with hybrid priors.
\newblock \emph{arXiv preprint arXiv:2407.12684}, 2024.

\bibitem[Zeng et~al.(2024)Zeng, Yang, Li, Liu, Zhang, Tian, Zhu, Guo, Wang, Xu, et~al.]{trans4d}
Bohan Zeng, Ling Yang, Siyu Li, Jiaming Liu, Zixiang Zhang, Juanxi Tian, Kaixin Zhu, Yongzhen Guo, Fu-Yun Wang, Minkai Xu, et~al.
\newblock Trans4d: Realistic geometry-aware transition for compositional text-to-4d synthesis.
\newblock \emph{arXiv preprint arXiv:2410.07155}, 2024.

\bibitem[Zhang et~al.(2025)Zhang, Chen, Wang, Liu, Wang, and Qiao]{4diffusion}
Haiyu Zhang, Xinyuan Chen, Yaohui Wang, Xihui Liu, Yunhong Wang, and Yu~Qiao.
\newblock 4diffusion: Multi-view video diffusion model for 4d generation.
\newblock \emph{Advances in Neural Information Processing Systems}, 37:\penalty0 15272--15295, 2025.

\bibitem[Zhang et~al.(2018)Zhang, Isola, Efros, Shechtman, and Wang]{lpips}
Richard Zhang, Phillip Isola, Alexei~A Efros, Eli Shechtman, and Oliver Wang.
\newblock The unreasonable effectiveness of deep features as a perceptual metric.
\newblock In \emph{Proceedings of the IEEE conference on computer vision and pattern recognition}, pages 586--595, 2018.

\bibitem[Zhang et~al.(2023)Zhang, Wang, Zhang, Zhao, Yuan, Qin, Wang, Zhao, and Zhou]{I2VGen}
Shiwei Zhang, Jiayu Wang, Yingya Zhang, Kang Zhao, Hangjie Yuan, Zhiwu Qin, Xiang Wang, Deli Zhao, and Jingren Zhou.
\newblock I2vgen-xl: High-quality image-to-video synthesis via cascaded diffusion models.
\newblock \emph{arXiv preprint arXiv:2311.04145}, 2023.

\bibitem[Zhao et~al.(2023)Zhao, Yan, Xie, Hong, Li, and Lee]{animate124}
Yuyang Zhao, Zhiwen Yan, Enze Xie, Lanqing Hong, Zhenguo Li, and Gim~Hee Lee.
\newblock Animate124: Animating one image to 4d dynamic scene.
\newblock \emph{arXiv preprint arXiv:2311.14603}, 2023.

\end{thebibliography}
}

\newpage
\appendix
\section{Videos of generated 4D NeRFs}
We provided a webpage with example objects and short videos of their 4D animations.
To view the content, please unzip the Supplementary.zip file first. Then, open the webpage.html file to explore the 4D object videos.

\section{view-consistency.}
\begin{figure}[h]
    \centering
    \includegraphics[width=0.75\linewidth]{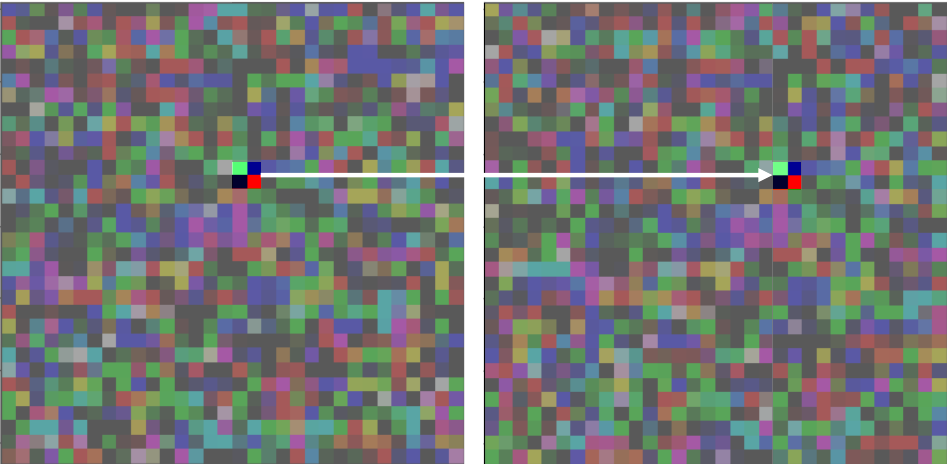}
    \caption{A specific noise patch on the sphere remains consistent across different camera viewpoints. The left and right panels show the same noise field rendered from two distinct viewpoints. The highlighted patch appears in different image locations due to the camera shift but retains identical structure and values. The white arrow indicates the 3D correspondence of the patch across views.} 
    \label{fig:view_consistent_noise_viz}
\end{figure}

Figure~\ref{fig:view_consistent_noise_viz} illustrates our viewpoint-consistent noising strategy. On the left, we visualize a fixed canonical noise field projected from one viewpoint, where the noise assigned to each pixel is derived from its corresponding face on a canonical 3D sphere. On the right, the same canonical noise field is reprojected from a different viewpoint. Crucially, the colored patch (highlighted in green, black, blue and red channels) remains consistent across views its position changes due to the viewpoint shift, but its local structure and values remain identical. The white arrow denotes the 3D reprojection path of the patch center between the two views. This example demonstrates how our method ensures consistent spatial alignment of noise across viewpoints, which helps preserve coherent appearance and motion cues throughout the distillation process.

\section{Sensitivity to prompt.}
\begin{figure}[h]
    \centering
    \includegraphics[width=0.92\linewidth]{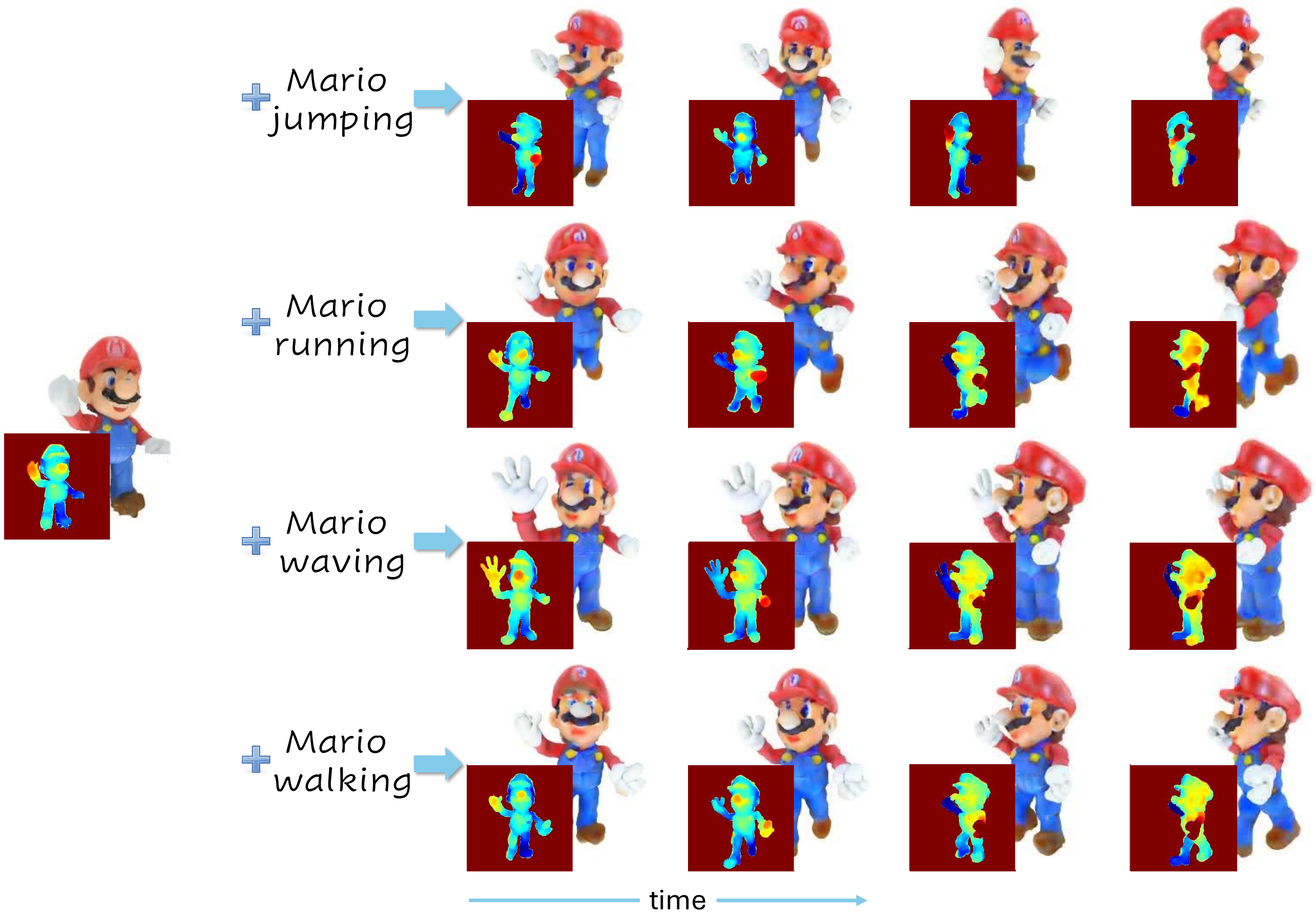}
    \caption{
    Different prompts generate different 4D, matching the movement description. The object in question is a Mario figure (on the left), and we provide three distinct prompts that describe three different dynamics of the figure. On the right, the generated 4D illustrates the corresponding movements based on these prompts. 
    } 
    \label{fig:prompt_effect}
\end{figure}

We explore the effect of different prompts, describing different dynamics, on the generated 4D scene. Figure.~\ref{fig:prompt_effect} shows the results when using the 3D object ``Mario" and supplying it with three different dynamic prompts: ``jumping" (top row), ``running" (middle row), and ``waving" (bottom row). The ``Mario" figure, moves differently, according to the specified actions in the description 

\section{Fail cases}
\begin{figure}[h]
    \centering
    \includegraphics[width=0.9\linewidth]{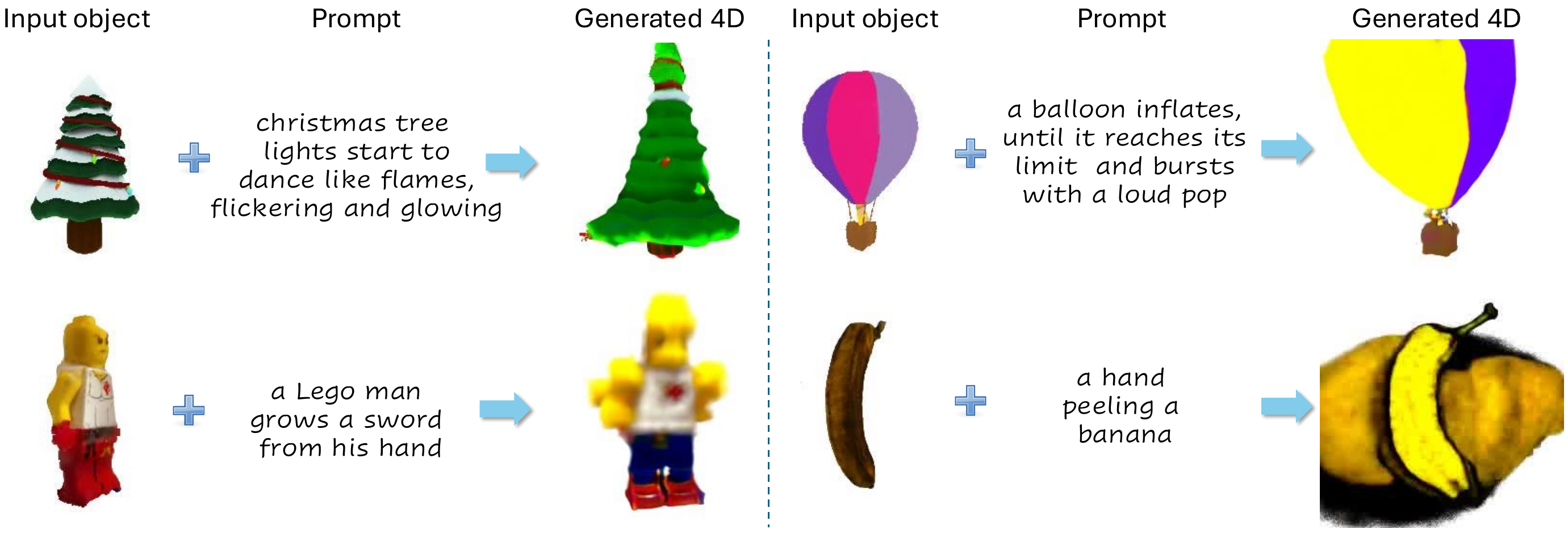}
    \caption{Despite plausible object-prompt pairings, the model occasionally fails to generate coherent or semantically aligned dynamics. These examples highlight limitations with \ourmethod{} prompt and dynamic depiction} 
    \label{fig:fails}
\end{figure}

Some object classes, particularly in the Objaverse dataset, cause severe deformation and color changes in the input object. These severe deformations cause the evaluation metrics to deviate significantly from the norm. Example shown in Figure~\ref{fig:fails}.
For example, the object classes we identified include "Christmas tree", "legoman", "sunflower", "banana" and "balloon"

\end{document}